\documentclass[journal]{IEEEtran}
\usepackage{amsmath,amsfonts}
\usepackage{algorithm}
\usepackage{array}
\usepackage[caption=false,font=normalsize,labelfont=sf,textfont=sf]{subfig}
\usepackage{textcomp}
\usepackage{stfloats}
\usepackage{url}
\usepackage{verbatim}
\usepackage{graphicx}
\usepackage{cite}
\usepackage{wasysym}
\usepackage{enumerate}
\usepackage{enumitem}
\usepackage[eulergreek]{sansmath}

\newcommand\sep{,\xspace}

\newif\ifieee\ieeefalse
\ieeetrue

\newif\ificga
\ifieee
\icgafalse
\else
\icgatrue
\fi

\usepackage{dcolumn}
\usepackage{comment}

\usepackage{subfig}
\usepackage{xspace} 
\usepackage{eqparbox}
\usepackage{xcolor}
\usepackage{array}
\usepackage{amsmath}
\usepackage{amsfonts}
\usepackage{amsthm}
\usepackage{makecell}
\usepackage{wasysym}
\usepackage{tikz}
\usepackage{pgfplots}
\pgfplotsset{compat=1.17}
\usetikzlibrary{shapes, arrows}
\usepackage{afterpage}

\newcolumntype{d}[1]{D{.}{.}{#1}}

\newcommand{\ORplayer}{\text{$\pi_{\lor}$}\xspace}
\newcommand{\ANDplayer}{\text{$\pi_{\land}$}\xspace}

\ificga
\firstpage{1}
\lastpage{6}
\volume{1}
\pubyear{2021}
\fi

\ificga
\renewcommand{\cite}{\citep}
\fi

\begin{document}

\ificga
\begin{frontmatter} 
\fi

%

\ificga
\runtitle{A Local-pattern Related Look-Up Table}
\fi

\ifieee
\title{A Local-Pattern Related Look-Up Table}
\fi

\ifieee
\author{Chung-Chin Shih, Ting Han Wei, Ti-Rong Wu, and I-Chen Wu, \IEEEmembership{Senior Member, IEEE}%
\thanks{This research was supported in part by the Ministry of Science and Technology (MOST) of the Republic of China (Taiwan) under Grant 108-2634-F-009-011, 109-2634-F-009-019, and 110-2634-F-009-022 through Pervasive Artiﬁcial Intelligence Research (PAIR) Labs, and the computing resource was supported in part by National Center for High-performance Computing (NCHC) of Taiwan. (Corresponding author: I-Chen Wu.)}

\thanks{Chung-Chin Shih is with the Department of Computer Science, National Yang Ming Chiao Tung University, Hsinchu, Taiwan (e-mail: rockmanray@aigames.nctu.edu.tw).}
\thanks{Ting Han Wei is with the Department of Computing Science, University of Alberta, Edmonton, Canada (e-mail: tinghan@ualberta.ca).}
\thanks{Ti-Rong Wu is with the Institute of Information Science, Academia Sinica, Taipei, Taiwan (e-mail: tirongwu@iis.sinica.edu.tw).}
\thanks{I-Chen Wu is with the Department of Computer Science, National Yang Ming Chiao Tung University, Hsinchu, Taiwan (e-mail: icwu@cs.nctu.edu.tw).}
}
\fi

\ifieee
\markboth{A Local-Pattern Related Look-Up Table}
{Shell \MakeLowercase{\textit{et al.}}: A Sample Article Using IEEEtran.cls for IEEE Journals}
\maketitle
\fi

\begin{abstract}
This paper describes a Relevance-Zone pattern table (RZT) that can be used to replace a traditional transposition table. 
An RZT stores exact game values for patterns that are discovered during a Relevance-Zone-Based Search (RZS), which is the current state-of-the-art in solving L\&D problems in Go. 
Positions that share the same pattern can reuse the same exact game value in the RZT.
The pattern matching scheme for RZTs is implemented using a radix tree, taking into consideration patterns with different shapes. 
To improve the efficiency of table lookups, we designed a heuristic that prevents redundant lookups. 
The heuristic can safely skip previously queried patterns for a given position, reducing the overhead to 10\% of the original cost.
We also analyze the time complexity of the RZT both theoretically and empirically. Experiments show the overhead of traversing the radix tree in practice during lookup remain flat logarithmically in relation to the number of entries stored in the table.
Experiments also show that the use of an RZT instead of a traditional transposition table significantly reduces the number of searched nodes on two data sets of $7 \times 7$ and $19 \times 19$ L\&D Go problems.
\end{abstract}

\ifieee
\fi

\ificga
\fi

\ificga
\end{frontmatter}
\fi

\section{Introduction}\label{sec:introduction}
\ifieee
\IEEEPARstart{A}{}
\else A
\fi
critical technique that many computer game programs use to boost search efficiency is the \emph{transposition table} (TT). 
Transpositions are positions that share a set of specific properties, most commonly defined to be the board configuration that is independent of move history. 
Using a lookup table, we can skip search for entire subtrees by replacing the value of positions in the search tree if its transposition has been encountered previously in the search.

The power of lookup tables can be significantly extended if value reuse can be extended beyond positions sharing the same board configuration. 
One such way to do this is to match local patterns on the board. 
The main obstacle here is ensuring correctness, i.e. whether boards with the same patterns truly share the same value. 
There have been many previous studies that use local patterns to suggest moves. 
For examples, to use $3 \times 3$ patterns to make moves in MCTS simulations \cite{munos2006modification}, to play cut in local patterns \cite{mueller1991pattern}, to use fixed size patterns to suggest strong moves \cite{boon1990pattern, cazenave2001generation, bayer2008gnu}. 
In Hex, \emph{virtual connections} \cite{hayward2003solving,hayward2006hex,henderson2009solving} are local patterns that can be identified to calculate the outcome of game positions. 
Many of these patterns involve handcrafted patterns as an offline database, or identifying specific patterns on the board as they are encountered during search, rather than storing the outcomes in a table and replacing positional values outright as in the case for transposition tables.

An example of a method that defines patterns automatically during search, which can then be stored and reused, is the so-called \emph{Relevance-Zone-Based Search} (\emph{RZS}) \cite{shih2022anovel}.
RZS was shown to be the state-of-the-art (SOTA) in solving what is commonly referred to as \emph{life-and-death} (\emph{L\&D}) problems in Go. 
In Go, safe (live) stone groups remain on the board indefinitely, and count towards the score of the player that owns it.
In contrast, unsafe (dead) stone groups do not count towards the final score since they can be captured and removed from the board if played out.
L\&D problems can be seen as a class of critical sub-problems that, when solved, contribute to a more accurate board evaluation in the full game of Go.
A \emph{relevance zone} (\emph{RZ}) is a mask on the game board that satisfies the property that any number of moves played outside of the zone do not affect the outcome of the position (L\&D in \cite{shih2022anovel}). 
This greatly reduces the search space since moves outside of the RZ do not need to be searched. 

\begin{figure}[t]
\centering
\subfloat[$p_a$]{\includegraphics[width=0.2\columnwidth]{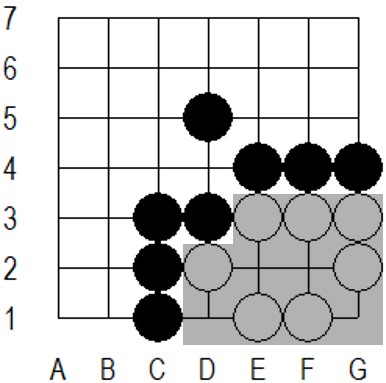}}
\subfloat[$p_{b'}$]{\includegraphics[width=0.2\columnwidth]{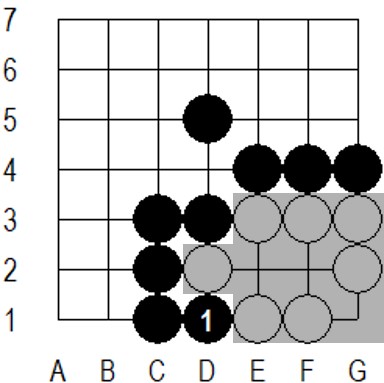}}
\subfloat[$p_{c'}$]{\includegraphics[width=0.2\columnwidth]{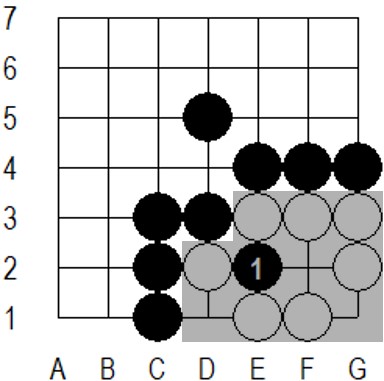}}
\subfloat[$p_{d'}$]{\includegraphics[width=0.2\columnwidth]{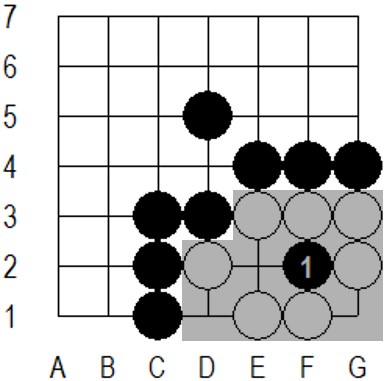}}
\\
\subfloat[$p_*$]{\includegraphics[width=0.2\columnwidth]{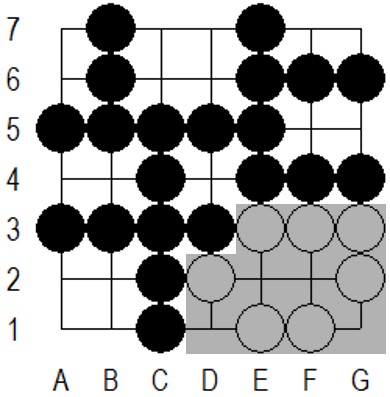}}
\subfloat[$p_b$]{\includegraphics[width=0.2\columnwidth]{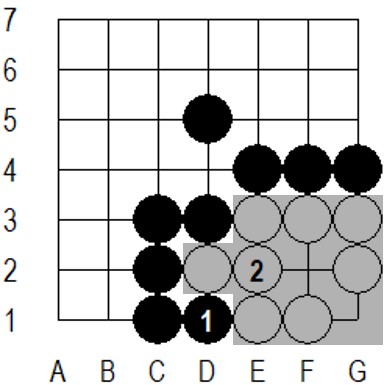}}
\subfloat[$p_c$]{\includegraphics[width=0.2\columnwidth]{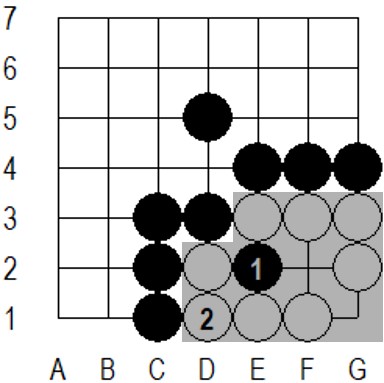}}
\subfloat[$p_d$]{\includegraphics[width=0.2\columnwidth]{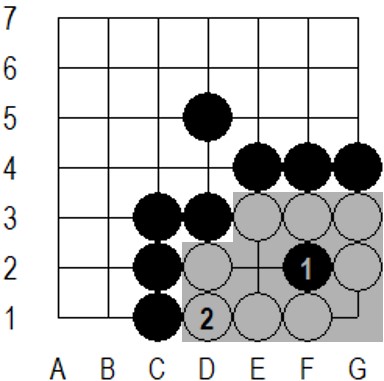}}
\caption{
An example to illustrate relevance-zones in Go.
}
\label{fig:examples}
\end{figure}

Fig. \ref{fig:examples} shows such an example for the game of $7 \times 7$ Kill-all Go. 
In $7 \times 7$ Kill-all Go, the rules are the same as standard Go, except Black plays two stones as its first move, and must capture the entire board to win. 
In contrast, White wins if any white stones remain alive on the board by the end of the game. 
Fig. \ref{fig:examples} (a) and (e) are two different positions, $p_a$ and $p_*$, that share the same RZ pattern, as shown in the shaded portion of the board. 
White wins in both positions simply by following the same strategy within the RZ. 
In fact, any arbitrary positions which share the same configuration inside the shaded region as in $p_a$ and $p_*$ will be a White win as shown in \cite{shih2022anovel}.

This paper focuses on the design of such a pattern-matching lookup table.
The lookup algorithm needs to locate, from a table of different zone patterns, any such pattern for any given position.
A naive method is by performing linear search, with a time complexity of $O(n)$, where $n$ is the number of the entries in the table. 
This is impractical as the number of zone patterns grows.
In comparison to previous efforts on using patterns as knowledge during search, the dynamically generated RZ patterns have various sizes and shapes.
In this paper, we propose a data structure that is based on radix trees (also known as Patricia trees) \cite{morrison1968patricia, mueller1991pattern}.
We also provide theoretical analysis of our method's time complexity. 
Empirical statistics are shown via experiments on $7 \times 7$ Kill-all Go and $19 \times 19$ L\&D problems to illustrate that the lookup times remain feasible as the search size increases.
The zone pattern matching mechanism, used as an auxiliary transposition table, significantly reduces the number of searched nodes.
Namely, the larger the number of search tree is, the more overhead the mechanism can reduce in general. 
This increases the number of solved L\&D problems from 68 \cite{shih2022anovel} to 83 out of a total of 106 L\&D problems within the same amount of computation time, about 500,000 search nodes.
Specifically, the number of searched nodes is reduced by a factor of about 4 on average for the two benchmarks used in the RZ paper \cite{shih2022anovel}. 

This article is organized as follows.
Section \ref{sec:background} defines \emph{life-and-death} (L\&D) and the terminology used for pattern matching.
We also review how traditional transposition tables are used and the baseline RZ algorithm.
Section \ref{sec:method} introduces our zone pattern matching mechanism, describes further improvements, and provides theoretical time complexity for such a lookup scheme.
Section \ref{sec:experiments} gives details on experiments that show how our method can improve the process of solving L\&D problems. 
Finally, we make concluding remarks in Section \ref{sec:conclusion}.

\section{Background}
\label{sec:background}

\subsection{Life-and-death Problems for Go}

The game of Go is played by two players, Black and White, who take turns placing stones of their own color on \emph{intersections} on the board.
Intersections are adjacent to each other in four directions; diagonals are not adjacent.
A set of stones of the same color that are connected via adjacency to one another is called a {\em block}.
Unoccupied intersections adjacent to a block are called {\em liberties} of that block. 
A block is captured if an opponent stone is placed in the block's last liberty.
Each block must have at least one liberty; players are not allowed to make suicidal moves that deprives the last liberty of their own blocks, unless that move also captures opponent stones.
{\em Safety} refers to a situation where blocks cannot be captured by the opponent under optimal play. 

While Go is a symmetric game where both players share the same goal, we present here an abstract asymmetric description from the perspective of one of these players. 
In a two-player game where one player tries to achieve a defined goal, and the other attempts to deny that goal, the former is called the \emph{OR-player}, denoted by \ORplayer, and the latter is called the \emph{AND-player}, denoted by \ANDplayer.
Achieving the goal in this context is equivalent to {\em winning} from the OR-player's perspective.

Life-and-death (L\&D) problems are sometimes referred to as tsumego.
They can be viewed as a subgame of Go where, given a position and some stones of \ORplayer designated as \emph{crucial} \cite{kishimoto2005search}, \ORplayer has the goal of guaranteeing safety for any of these crucial stones. \ANDplayer will try to capture all crucial stones. 

A set of blocks is said to be {\em unconditionally alive (UCA)} \cite{benson1976life} if the opponent cannot capture it even if the owner of the block always passes. In Fig. \ref{fig:examples}, the white blocks of $p_b$, $p_c$ and $p_d$ are UCA.
Achieving UCA for any crucial stones is a valid solution to a L\&D problem.\footnote{For simplicity, this paper does not consider ko or super-ko, i.e. we allow repetitions. Player \ORplayer only wins if they can achieve safety without engaging in any ko fights, as \cite{shih2022anovel} did.}

A specific instance of a L\&D problem is the so-called $7 \times 7$ Kill-all Go game, in which Black, the first player, plays two stones initially, then both players (starting with White) play one stone per turn alternately on a $7 \times 7$ Go board. 
White, playing as \ORplayer, must achieve safety for at least one white block via UCA, while Black, playing as \ANDplayer, must kill all white stones.
We use both $7 \times 7$ Kill-all Go and $19 \times 19$ L\&D problems in our experiments.

\subsection{Transposition Tables}
\emph{Transposition tables} (\emph{TT}) store heuristic and exact evaluations of searched positions for future reuse.
Zobrist hashing \cite{zobrist1990new} is often used to differentiate between transpositions. 
A simple transposition scheme is one where positions that share the same stone configurations and player to play belong in the same category. 
A traditional TT lookup can be implemented in various ways, e.g. open addressing hash tables \cite{knuth1997art} or binary search trees \cite{windley1960trees}.

Furthermore, there are some games where we need to differentiate between positions that have different move histories. 
E.g. when considering the castling rule in chess, the evaluation may be different for positions with the same board configurations and the player to play. The legality of moving king and castle pieces depends on whether they have moved before.
For Go, ko and super-ko rules prohibit players from repeating previously encountered positions.
In cases where the board appears to be the same, but certain moves are prohibited due to the super-ko rule, extra care must be taken so that incorrect values are not used.
This is the so-called \emph{Graph-History-Interaction} (\emph{GHI}) Problem \cite{palay1983searching}.
Many algorithms have been proposed to deal with GHI \cite{breuker2001solution,kishimoto2004general,kishimoto2005solution}; we do not consider repetitions in this paper.

\subsection{Pattern Matching}
\label{subsec:definition}

Given a position $p$, $\pi(p)$ is the player to play, and $\beta(p)$ is the specific configuration of black stones, white stones, and empty intersections that make up the position.
A subset of intersections $z$ is called a \emph{zone}, and $\beta(p)\odot z$ is the configuration of stones on the position that lies within $z$, which we call a {\em zone pattern}.

Formally, a zone pattern is specified as $\phi=(\pi_\phi, z_\phi, \beta_\phi$), where $\pi_\phi$ is the player to play, $z_\phi$ is a zone that the pattern is located in, and $\beta_\phi$ is the stone configuration in $z_\phi$. 
Since black and white stones outside of $z_\phi$ are not part of the pattern, i.e. they are not considered during pattern matching, no information outside of $z_\phi$ are stored into $\beta_\phi$.
Let us assume that two positions $p$ and $q$ are different, i.e. $\beta(p) \neq \beta(q)$, and that 
there exists a zone pattern $\phi$ such that $\beta(p) \odot z_\phi = \beta(q) \odot z_\phi = \beta_\phi$, and that it is $\pi_\phi$'s turn to play for $p$ and $q$. 
Under these circumstances, we say that $p$ and $q$ \emph{match} the zone pattern $\phi$.

It is worth noting that pattern matching between patterns and positions is a many-to-many relation. 
That is, there may be many positions with the same zone pattern, while there also may be many different matching zone patterns for a position.
This makes zone pattern matching potentially more complicated than a traditional TT lookup.

\subsection{Relevance-Zone-Based Search}
\label{subsec:rzone}

A zone $z$ can be called a relevance-zone (RZ) with respect to a winning position $p$, if the following property is satisfied \cite{shih2022anovel}.
\begin{description}
\item[RZ-P] 
For all positions $p_*$ with $\beta(p_*) \odot z = \beta(p)\odot z$ and $\pi(p_*)=\pi(p)$, $p_*$ are also wins (for $\ORplayer$).
\end{description}

For Go and Kill-all Go, there will always exist at least one RZ with respect to a winning position. 
As an example, in the most extreme case, the RZ can contain the entire board; this satisfies RZ-P.
In the case that UCA is achieved in $p$, an RZ satisfying RZ-P includes all intersections of the unconditionally safe blocks and their regions \cite{benson1976life}, since these blocks are safe regardless of any moves played outside the RZ by the opponent.

Relevance-Zone-Based Search (RZS) \cite{shih2022anovel} is a search algorithm that takes advantage of RZs.
An example as shown in Fig. \ref{fig:examples} (a) is used to illustrate RZS.
For position $p_a$, if Black plays at 1 as in $p_{b'}$, $p_{c'}$, and $p_{d'}$, White can reply 2 to win as in $p_b$, $p_c$, and $p_d$, respectively. 
Since positions $p_b$, $p_c$, and $p_d$ reach UCA for white stones, we can designate the RZs as the shaded intersections, which we denoted by $z_b$, $z_c$, and $z_d$, respectively. 
For their previous positions $p_{b'}$, $p_{c'}$, and $p_{d'}$, White wins by playing 2 and therefore their corresponding RZs (satisfying RZ-P) are also $z_b$, $z_c$, and $z_d$, respectively. 
The patterns inside these zones form zone patterns called \emph{RZ patterns}. 
Since Black 1 in Fig. \ref{fig:examples} (f) is outside of the RZ $z_b$ satisfying RZ-P, it can be viewed as a null move. 
From RZ-P, White wins by replying at 2 for all Black moves outside $z_b$. 
Thus, White wins for all Black moves. 
Let the zone $z_a$ be a union of $z_b$, $z_c$, and $z_d$, as the shaded in Fig. \ref{fig:examples} (a).
Actually, $z_a$ is an RZ satisfying RZ-P, since White wins for all positions $p_*$ matching $z_a$ by replaying the above. 
A proof for the correctness of replaying the strategy are in \cite{shih2022anovel}, omitted in this paper.

In \cite{shih2022anovel}, their experiments for RZS were based on a modified \emph{Monte-Carlo Tree Search} (\emph{MCTS}) \cite{coulom2006efficient} algorithm that propagates win/loss values towards the root. 
Following this mechanism, RZS can greatly reduce the search space. 
RZS was a SOTA in solving L\&D problems \cite{shih2022anovel} then, solving 68 among 106 L\&D problems in a L\&D book, written by Cho Chikun \cite{cho1987life}, a Go grandmaster.

\section{Our Method}
\label{sec:method}

This section describes the details about our method.
Subsection \ref{subsec:difference} first discusses the difference between past study and our method.
Subsection \ref{subsec:tree} describes the data structures used and the zone pattern table. Subsection \ref{subsec:enhancements} lists additional enhancements that improve the base method. 
Finally, we analyze the time complexity of our method in subsection \ref{subsec:complexity}.

\subsection{Previous Pattern Matching Methods}
\label{subsec:difference}

Previous methods on pattern matching often build an offline database of handcrafted patterns with expert domain knowledge \cite{munos2006modification, mueller1991pattern, boon1990pattern, cazenave2001generation, bayer2008gnu}.
Such methods are more interested in recognizing local patterns which help identify urgent situations.
These local patterns are often identified throughout the whole board from a static database, which does not change during search.

In RZS, there exist many (at least one) RZ for each identified winning position. 
RZ patterns corresponding to these RZs can be saved and their outcomes can be reused in other searched positions. 
Since new RZ patterns are generated throughout the RZS algorithm, a growing set of RZ patterns are inserted into a table, which we call an \emph{RZ pattern table} (\emph{RZT}).

Another major difference from previous pattern matching methods is that the shapes of the zones in previous pattern matching techniques tend to be fixed and contiguous \cite{mueller1991pattern}.
The zone shapes in our pattern matching mechanism are dynamically decided through frequency analysis. 
Thus the zone shapes do not need to be contiguous.

\subsection{Pattern Matching Radix Tree}
\label{subsec:tree}

For better efficiency, the zone pattern table can be stored as a tree. 
To do so, we first define an ordered set of relevant intersections on the game board. 
These so-called \emph{ordered crucial intersections} (\emph{OCI}), $(I_1,I_2,...,I_n)$, can be used to classify zone patterns $\phi$. 
By mapping the contents of each crucial intersection to predefined characters, the set of OCI for $p$ can be expressed as a string of length $n$. 
We can then construct a radix tree \cite{morrison1968patricia}, which can be used to finding a matching $\phi$ for $p$ efficiently.

\begin{figure}[t]
\centering
\subfloat[]{\includegraphics[width=1.0\columnwidth]{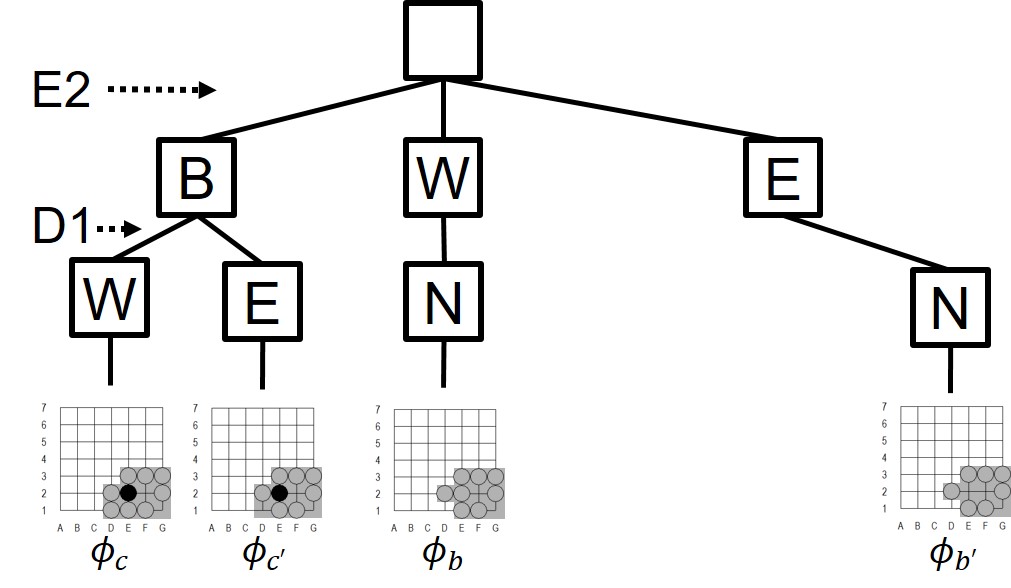}}
\\
\subfloat[]{\includegraphics[width=1.0\columnwidth]{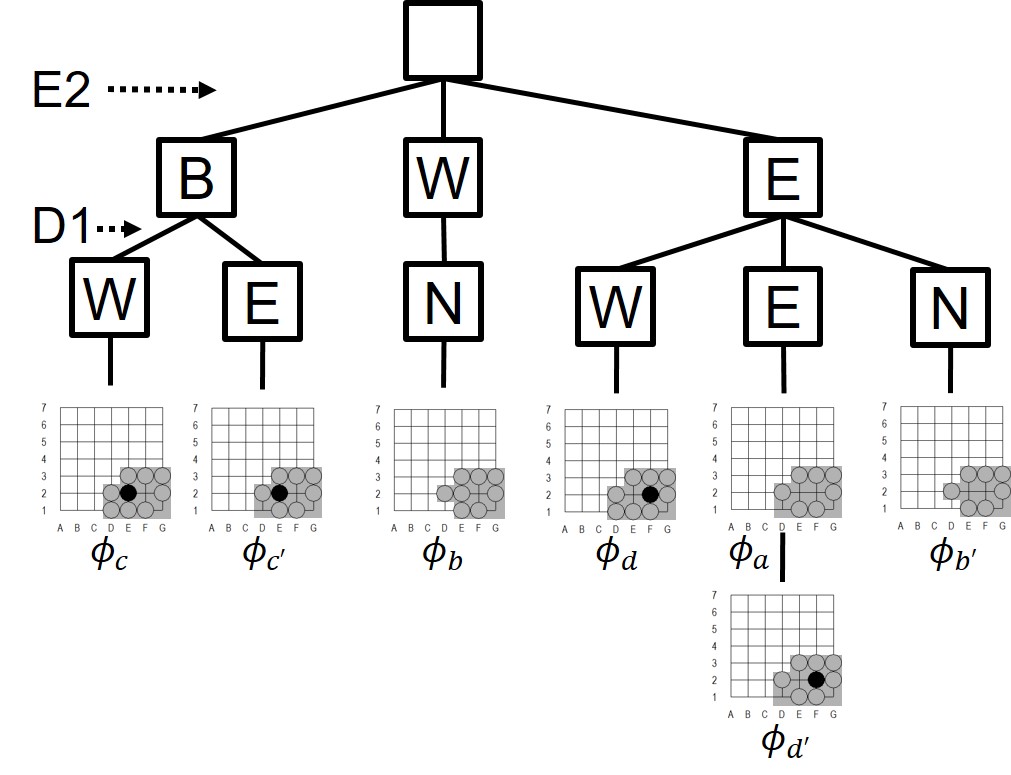}}
\caption{ (a) The radix tree of the table built upon the OCI (E2,D1) among the four RZ patterns $\phi_b$, $\phi_{b'}$, $\phi_c$, and $\phi_{c'}$ in Fig. \ref{fig:examples}. (b) The radix tree of the table built upon the OCIs (E2,D1) among all the RZ patterns in Fig. \ref{fig:examples}.}
\label{fig:oci_fig}
\end{figure}

More specifically, at depth $d$ of the tree, the lookup is directed to one of four branches depending on what is located on $I_d$: $B$ for a black stone, $W$ for a white stone, $E$ for an empty intersection, and $N$ for intersections that are not matched (outside the zone $z_\phi$), i.e. "don't care".
After querying all OCI, a linked list is used to store all zone patterns that share the same ordered crucial string.
For example, there are seven different zone patterns in Fig. \ref{fig:examples}. 
Suppose the OCI $(I_1, I_2)$ are (E2, D1).
The table's resulting radix tree with seven zone patterns is shown in Fig. \ref{fig:oci_fig} (b), where both $\phi_a$ and $\phi_d$ have the same OCI and therefore are linked as a list.
Note that the tree can be separated into two in the real implementation, one for those Black to play and the other for White. 

\subsubsection{Zone Pattern Insertion}
\label{subsubsec:insertion}

When a zone pattern $\phi=(\pi_\phi, z_\phi, \beta_\phi$) needs to be inserted into the table, we traverse the tree as in the case with string matching using a radix tree. 
That is, at each depth $d$ of the tree, if $I_d \in z_\phi$, we encode the $d$th character of the pattern string with $B$, $W$, or $E$, if the intersection is occupied by Black, White, or empty, respectively; if $I_d \notin z_\phi$, the character is encoded as $N$.
Thus, each character in the pattern string corresponds to a choice of one of four branching child nodes in the radix tree.
Once all $n$ OCI are encoded into a string, we traverse the tree and append $\phi$ to the tail of the linked list in the corresponding leaf node.

We illustrate an example as in Fig. \ref{fig:examples}.
After RZS searches to position $p_{b}$ with a win, a zone pattern $\phi_b$ is stored into the radix tree. 
When RZS is back to $p_{b'}$ (with a win too), another zone pattern $\phi_{b'}$ is inserted. 
Similarly, after RZS search to $p_c$ and is back to $p_{c'}$, both $\phi_{c}$ and $\phi_{c'}$ are inserted. 
Then, the radix tree is shown in Fig. \ref{fig:oci_fig} (a).
After finishing the search to $p_d$, $p_{d'}$ and $p_a$, the radix tree is shown in Fig. \ref{fig:oci_fig} (b). 
Particularly, when inserting $\phi_b$ and $\phi_{b'}$ where D1 is outside the zone pattern, they should be inserted into node $N$. 

\subsubsection{Zone Pattern Matching}
\label{subsubsec:search}

The process of searching for a zone pattern $\phi$ that matches an input position $p$ works as a backtracking search algorithm.
At each depth $d$ of the radix tree, we traverse to one of the three children at $I_d$, namely B, W, and E, according to what kind of stone is placed on $I_d$ in $p$.
Since arbitrary zone patterns may contain any number of intersections (up to the whole board), patterns that belong in the same linked list share only partial matches $I_1, ..., I_n$. 
Therefore, once we reach a leaf node, a linear search is performed to find any matching zone patterns for $p$.
If no matching patterns are found in the linked list, we return to previous depths of the radix tree and traverse down the $N$ child, if there's any pattern below it.

We use the process of matching a pattern for $p_*$ in Fig. \ref{fig:examples} (e) as an example, using the RZT in Fig. \ref{fig:oci_fig} (b).
Suppose it is Black's turn.
By following two successive intersections at E2 and D1 to match $\phi_a$, which is in Black's turn too, we can conclude that $p_*$ is a win.
Let us consider another position $p_*^{w}$ with the same board configuration as $p_*$, but White to play.
Despite the matching intersections, the player to play is now different, so we cannot match $\phi_a$ (even if it is actually a win in this case).
We must backtrack and search along the node $N$, matching with $\phi_{b'}$ for $p_*^{w}$.

\subsubsection{Reconstructing the Radix Tree}
\label{subsubsec:reconstruction}

The choice of the set of OCI affects insertion and search efficiency. 
If a different OCI set is likely to improve efficiency, it may be necessary to reconstruct the radix tree accordingly.
This can be done simply by collecting all zone patterns, then inserting them into a new radix tree using the new OCI set.

We design the following heuristic reconstruction scheme in an attempt to minimize the reconstruction cost while maintaining efficiency over the lifetime of the zone pattern table. 
For the first 1000 entries in the table, we reconstruct the tree every 100 entries.
Note that for the first 100 entries, the tree of the table is actually a linear list.
Afterwards, the radix tree is reconstructed whenever the table size has increased by a factor of 10\%.
Note that if the new OCIs are the same as the old ones, the reconstruction is not needed.
More enhancements are described in subsection \ref{subsec:enhancements}.

\subsection{Enhancements}
\label{subsec:enhancements}


\subsubsection{Dynamically Determining the Elements of the OCI Set}
\label{subsubsec:heapmap}
There are many ways to determine the OCI set in order to maintain efficiency.
One general way is to count the occurrence of intersections among all patterns in the table.
The tally of intersection occurrences can be sorted using a heap.
The heap is initialized with the same number of nodes as the board size (total number of intersections on the board). 
When a zone pattern is found, we keep track of the tally for each intersection that is included in the zone. 
Whenever the radix tree is reconstructed, we order heap by each intersection's occurrence count from largest to smallest.
The top 80\% are then chosen to form the new OCI set $(I_1,I_2,...,I_n)$, and subsequently used for reconstruction.

\begin{figure}[t]
\centering
\includegraphics[width=0.8\columnwidth]{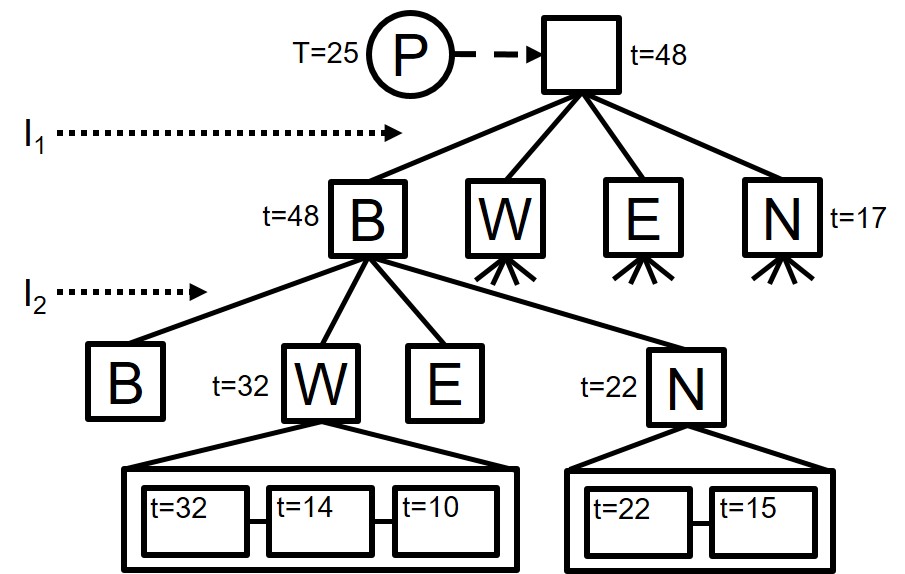}
\caption{A radix tree with timestamps.}
\label{fig:tree_examples}
\end{figure}

\subsubsection{Reducing Wasteful Table Lookups}
\label{subsubsec:timestamp}

RZ matching is powerful in that a match can be used to skip entire subtrees during RZS search via strategy reuse.
However, frequent pattern table queries can become costly as the number of zone patterns increases.
This overhead is especially wasteful if the successful lookup rate is low, however, we still need to query a zone pattern matching whenever we encounter a position. 
Namely, in our experiments, there are about 2\% hit rate and low lookup success rates are somewhat expected, as described as follows.
In the case of a successful lookup on a position, the outcome is that the position is a win, i.e. there exists a replayable strategy to win. 
Thus, in the RZS tree (MCTS in \cite{shih2022anovel}), the search path to the position is traversed no longer. 
However, in case of a lookup failure on the position, the outcome is unknown, and therefore it is still highly likely to traverse to the position again along the RZS path.
Given the low hit rate, techniques that reduce failed table lookups are worth investigating.

To reduce the cost and number of failed table lookups, we introduce the concept of timestamps to skip unnecessary lookups during RZS, particularly when no changes (or small changes) have been made to the pattern table since a position was last queried.
Specifically, in the process of pattern insertion, we maintain a \emph{global timestamp} to be the current total number of entries in the table. 
When inserting a pattern $\phi$ into the table, let the timestamp $t$ of each node on the path of the radix tree be the global timestamp.
This ensures that the timestamps for all nodes represent the latest timestamps of inserting any of descendants of the nodes. 

Next, for each RZS node with position $p$, we store a different timestamp $T$ (initialized as zero) which is updated to the global timestamp when the latest failure lookup on $p$.
From above, we have the following important property. 
During table lookup, if $T\geq t$ for any traversed node with timestamps $t$ in a radix tree or subtree, we can guarantee that the lookup will be unsuccessful since no new pattern entries have been inserted in the subtree since both $t$ and $T$.
This means the lookup can end prematurely, saving the cost of traversing the radix tree.


An illustration is shown in Fig. \ref{fig:tree_examples}. 
The position $p$ with $T=25$ indicates that the time was $25$ when $p$ was last queried unsuccessfully.
The timestamp of the root of the radix tree is $t=48>T$. 
Since the radix tree may be updated after 24 and before 48, we need to check the tree. 
Suppose that the intersection at $I_1$ on $p$ is black. 
Then, we traverse to the node B and check if $T \geq t$. 
It is not ($25<48$), so we continue along this node.
Again, suppose that the intersection at $I_2$ on $p$ is white. 
Then, we traverse to the node $W$, a leaf node, and continue since $T < t$ ($25<32$). 
A linear search is performed from the head of the linked list. 
We search towards the tail of the list for a match to $p$. 
When the timestamp $t$ of a pattern is smaller than $T$, we can terminate the linear search prematurely. 
In this example, this happens at the node with $14$, since $14<25$.
Let us say in this example no patterns in the linked list provided a match. 
We backtrack to the node $N$ and immediately stop the table lookup, since $T \geq t$ ($25 \geq 22$), i.e., a failed lookup for $p$ was performed at $25$ after an insertion into the subtree rooted at the $N$ ($t=22$). 

Now, let the whole lookup on $p$ fail. 
Then, the timestamp on $p$ becomes $T=48$. 
Thus, next lookups on $p$ will not be traversed as long as the table remains unchanged. 
Namely, since $t=48$ remains in this situation, $T \geq t = 48$.
This is very helpful if RZS is based on MCTS, where each simulation always runs from the root to leaves. 
Those positions close to the root are visited and looked up many times, so their timestamps are likely to be high and thus tend to save a large amount of lookup cost. 


During reconstruction, a new set of OCI will be defined, changing the structure of the tree.
Each entry in the old tree will be inserted into the newly reconstructed tree in the same order, ensuring that the timestamps will update correctly.

\subsection{Time Complexity}
\label{subsec:complexity}
The data structure of our method is based on a radix tree, with the only difference being that the values (or RZ patterns) are stored in the leaf node with a linked list. 
The time complexity of a lookup on a radix tree is $O(h)$ where $h$ is the height of the tree. 
In our scenario, however, in the case that a matching pattern cannot be found, we must also backtrack and traverse down the node $N$ to look up for other possible matches.
Assuming the keys are uniformly distributed, the recurrence time complexity for lookups is $T(n) = 2T(n/4) + O(1)$ where $n$ is the number of nodes in the tree. 
This can be inferred as $T(n)=n^{0.5}$ following the Akra-Bazzi solution \cite{leighton1996notes}.
In practice, the OCIs are determined according to the occurrences of each intersection, sorted from more frequent to less frequent.
Therefore, fewer nodes are inserted under the $N$ branches.
Assuming only one-ninth of the keys are stored in the $N$ branches and no more than one-third in each of the other three branches, the recurrence relation is $T(n) = T(n/3) + T(n/9) + O(1)$, $T(n)\approx n^{0.44}$.

\begin{table}[th]
\centering
\caption{$7 \times 7$ L\&D Problems} \label{tab:exp_7x7}
\begin{tabular}{|c|c|c|}
\hline
 & RZS-TT & RZT-OCI\\
\hline
Solved Problems & 20/20 & 20/20\\
\hline
Simulations & 3,166,933 & 692,338\\
\hline
Time & 4596.512 & 1013.592\\
\hline
\end{tabular}
\end{table}

\begin{table}[th]
\centering
\caption{$19 \times 19$ L\&D Problems} \label{tab:exp_19x19}
\begin{tabular}{|c|c|c|}
\hline
 & RZS-TT & RZT-OCI\\
\hline
Solved Problems & 68/106 & 83/106\\
\hline
Simulations (Both 83 problems) & 10,724,163 & 2,264,406\\
\hline
Time (Both 83 problems) & 14188.882 & 2825.218\\
\hline
\end{tabular}
\end{table}

\section{Experiments}
\label{sec:experiments}

\subsection{Solving 7x7 and 19x19 L\&D Problems}
\label{subsec:solving_problems}

To verify the correctness and measure the performance of our RZT, we try to solve a set of problems for $7 \times 7$ Kill-all Go and $19 \times 19$ Go. 
The solver program is an AlphaZero MCTS solver \cite{silver2018general,winands2008monte} that is modified from the RZS program \cite{github_url,shih2022anovel}, which in turn is based on the Go player CGI \cite{wu2020accelerating}.
As an AlphaZero program, the solver's neural network components are trained using game samples generated through self-play.
One major difference between the solver's neural networks and a standard AlphaZero-style program is that it is trained using the so-called Faster to Life (FTL) method, also similar to proof-cost network in \cite{wu2021pcn}.
The FTL method modifies the winning conditions for self-play so that the training agents prefer to win with the fewest number of moves. 

All experiments are conducted on a machine with an Intel 2.5 GHz E5-2678 CPU, 128GB memory, and a GeForce GTX 1080-Ti GPU.
We use the same data set as \cite{shih2022anovel}, consisting of 20 problems for $7 \times 7$ Kill-all Go and 106 problems for $19 \times 19$ Go.
The $19 \times 19$ Go problems are selected from a well-known collection named the {\em "Life and Death Dictionary"}, written by Go grandmaster Cho Chikun \cite{cho1987life}. 

We use two versions for comparison, RZS with TT, called RZS-TT, and RZT with dynamic OCI, called RZT-OCI. 
For each of $7 \times 7$ problems, a maximum of 500,000 simulations are given to solve. 
All $7 \times 7$ problems can be solved by both versions, while RZT-OCI is more efficient in simulations by a factor of about 4.57 ($\approx$ 3,166,933/692,338).
For each of $19 \times 19$ problems, a time limit of 5 minutes is given to solve.
RZT-OCI is able to solve 15 more problems than RZS-TT (using TT) in this problem set.
We next compare the total number of simulations (nodes searched) by allowing unlimited time for the problems that are solvable with RZT-OCI but not with RZS-TT.
RZT-OCI is more efficient in simulations by a factor of about 4.74 ($\approx$ 10,724,163/2,264,406).
The details including the number of simulations, time, number of table entries, etc. for each problem are listed in Tables \ref{tab:exp_data_7} and \ref{tab:exp_data_19}.


\subsection{Analysis on Different Settings}
\label{subsec:traverse_cost}

\begin{figure}[t]
\begin{tikzpicture}
\begin{axis}[
    height=7cm,
    width=9cm,
    xlabel={106 19x19 Problems},
    ylabel={log scale (base 2) for traversing cost},
    xmin=0, xmax=106,
    ymin=0, ymax=40,
    xtick={},
    xticklabels={},
    ytick={5, 10, 15, 20, 25, 30, 35, 40},
    legend pos=south east,
]

\addplot[
    line width=1.5pt,
    color=orange,
    mark=square,
    mark size=1pt,
    error bars/.cd,
    y dir = both,
    y explicit
    ]
    coordinates {
    (1,9.285402219) (2,14.19752388) (3,11.39339046) (4,12.20853921) (5,21.22007656) (6,15.711425) (7,17.24590546) (8,15.52790457) (9,21.89318079) (10,18.19201824) (11,18.34351106) (12,17.08703804) (13,20.00881294) (14,21.33323071) (15,17.70686021) (16,20.59841331) (17,19.52045947) (18,21.51823197) (19,19.51205756) (20,20.12509559) (21,20.91167725) (22,18.1973367) (23,20.27512875) (24,25.0954621) (25,20.1429646) (26,21.12901946) (27,20.74293766) (28,21.82789778) (29,21.19943931) (30,22.43959326) (31,23.66884766) (32,21.57036902) (33,23.2971189) (34,24.11639695) (35,20.88858645) (36,23.72040689) (37,22.87747942) (38,22.75234083) (39,21.30189804) (40,22.72882834) (41,20.05961512) (42,23.3401727) (43,21.91666054) (44,21.61555825) (45,22.74984631) (46,20.71001267) (47,24.73217597) (48,22.07572526) (49,22.89878127) (50,23.28807554) (51,24.10452621) (52,25.35631605) (53,23.55360809) (54,22.60427609) (55,23.69765716) (56,25.02949547) (57,26.36700903) (58,24.87374598) (59,24.3985889) (60,24.16348853) (61,26.30748691) (62,33.71426837) (63,25.16359682) (64,25.79550556) (65,27.0915359) (66,25.27261808) (67,26.14689689) (68,27.74877695) (69,29.91879652) (70,25.77909131) (71,29.52712914) (72,27.54926263) (73,34.93049231) (74,27.21002389) (75,26.49829051) (76,26.80223048) (77,27.02145359) (78,26.36199486) (79,31.88863292) (80,32.61841654) (81,33.01752826) (82,31.16578916) (83,26.30500003) (84,28.35550714) (85,33.51949038) (86,35.29807255) (87,31.4076734) (88,32.80189014) (89,28.54991038) (90,36.68304796) (91,30.25674925) (92,28.81039306) (93,28.73775423) (94,31.8483542) (95,33.64048349) (96,34.86837616) (97,31.07108237) (98,33.63343906) (99,31.08824915) (100,31.71210806) (101,31.59689994) (102,35.38629683) (103,32.98378656) (104,34.25836202) (105,34.4032602) (106,32.64313393)
    };
    \addlegendentry{{LR-NT}}

\addplot[
    line width=1.5pt,
    color=black,
    mark=o,
    mark size=1pt,
    error bars/.cd,
    y dir = both,
    y explicit
    ]
    coordinates {
    (1,8.764871591) (2,9.471675214) (3,10.74315139) (4,11.53332973) (5,12.41785251) (6,13.65865779) (7,14.62684976) (8,14.71799765) (9,15.57981679) (10,16.16427844) (11,16.46585333) (12,16.47227833) (13,16.81262858) (14,16.97053285) (15,17.00478007) (16,17.16535848) (17,17.21948052) (18,17.21990587) (19,17.22088841) (20,17.25830839) (21,17.5772149) (22,17.85646808) (23,18.04470895) (24,18.10130376) (25,18.22010904) (26,18.29550888) (27,18.32268517) (28,18.32670597) (29,18.57278903) (30,18.65913533) (31,18.68288146) (32,18.80078301) (33,18.88918309) (34,19.2978074) (35,19.34495685) (36,19.40448051) (37,19.56061458) (38,19.63094607) (39,19.63217519) (40,19.66946123) (41,19.79176384) (42,20.06828364) (43,20.10493014) (44,20.18012891) (45,20.23029609) (46,20.32688807) (47,20.34113873) (48,20.34518106) (49,20.41652748) (50,20.53085779) (51,20.84879407) (52,20.90999487) (53,20.96648996) (54,21.08019881) (55,21.13846681) (56,21.19361138) (57,21.32929707) (58,21.38777018) (59,21.44648614) (60,21.7379182) (61,22.38666612) (62,22.45634464) (63,22.94458553) (64,22.97218839) (65,23.04218388) (66,23.15760398) (67,23.23774855) (68,23.47894695) (69,23.67104191) (70,23.94434621) (71,24.27419528) (72,24.38684929) (73,24.39111304) (74,24.43522826) (75,24.46033128) (76,24.54827525) (77,24.55712722) (78,24.70528266) (79,24.76244485) (80,25.35546704) (81,25.35684424) (82,25.72879604) (83,25.97394025) (84,26.34383724) (85,26.41055718) (86,26.74632044) (87,27.33376415) (88,27.39499121) (89,27.44484714) (90,27.44572739) (91,27.4509764) (92,27.50770916) (93,27.65893886) (94,27.71861934) (95,27.85786504) (96,27.93924552) (97,28.05066701) (98,28.11465557) (99,28.21242064) (100,28.24216246) (101,28.28810153) (102,28.47611059) (103,28.60232563) (104,28.7115287) (105,29.09039735) (106,29.60720262)
    };
    \addlegendentry{{LR-TS}}
    
\addplot[
    line width=1.5pt,
    color=red,
    mark=+,
    mark size=1pt,
    error bars/.cd,
    y dir = both,
    y explicit
    ]
    coordinates {
    (1,16.01321361) (2,14.99072446) (3,12.67639789) (4,12.0164601) (5,19.99839072) (6,16.62343858) (7,17.67618437) (8,21.71680188) (9,17.33322525) (10,17.77225013) (11,21.58016792) (12,17.91120471) (13,19.32453437) (14,22.65771685) (15,17.55783013) (16,17.38967335) (17,18.40332831) (18,20.21997675) (19,18.74588109) (20,21.01605605) (21,19.28273779) (22,17.61657756) (23,22.30717523) (24,22.1916394) (25,19.72326264) (26,19.53431998) (27,18.7041029) (28,18.35917045) (29,19.05978805) (30,18.13561911) (31,22.96195815) (32,19.78456616) (33,21.52382897) (34,22.27970579) (35,20.20893803) (36,22.29746179) (37,21.26831583) (38,18.46448592) (39,19.61837116) (40,19.54706585) (41,19.88213104) (42,20.01353532) (43,19.10718897) (44,18.88778287) (45,21.85896255) (46,22.17548651) (47,23.50968248) (48,23.843444) (49,23.22849755) (50,19.71538056) (51,21.97641344) (52,22.2112847) (53,20.12549041) (54,22.87882776) (55,20.16997025) (56,21.41321832) (57,19.45805811) (58,21.37017974) (59,23.48003787) (60,23.14061652) (61,21.30591924) (62,27.10384163) (63,21.1324627) (64,22.34643393) (65,26.8760249) (66,21.96959696) (67,21.63132629) (68,22.0133755) (69,24.23847574) (70,24.86569618) (71,26.75698786) (72,25.25197945) (73,26.50715872) (74,26.47983169) (75,23.19900084) (76,27.47393482) (77,24.45578354) (78,24.10143594) (79,24.24515133) (80,26.8106358) (81,25.38383812) (82,26.9030919) (83,27.85718873) (84,25.34082372) (85,24.53410509) (86,26.36921113) (87,27.01579316) (88,25.81459127) (89,26.88162529) (90,25.62447499) (91,24.82900641) (92,24.31636682) (93,23.42239269) (94,26.35647796) (95,26.4590665) (96,27.80307137) (97,25.23711855) (98,26.51829559) (99,28.18522029) (100,23.88243157) (101,25.57254741) (102,27.50149336) (103,24.61902712) (104,26.81387515) (105,26.64432165) (106,24.93122165)
    };
    \addlegendentry{{TR-NT}}
    
\addplot[
    line width=1pt,
    color=blue,
    mark=x,
    mark size=1.5pt,
    error bars/.cd,
    y dir = both,
    y explicit
    ]
    coordinates {
    (1,9.216745858) (2,12.96036401) (3,9.932214752) (4,9.695228291) (5,16.2649923) (6,13.58320032) (7,16.06137119) (8,13.31387503) (9,15.83540897) (10,15.52481715) (11,20.11844687) (12,17.76112916) (13,17.31582178) (14,18.54198139) (15,16.70253797) (16,16.97917954) (17,17.32091511) (18,17.27476942) (19,16.74644871) (20,17.48093217) (21,16.14427984) (22,15.31490183) (23,17.12093645) (24,19.34828079) (25,18.0647796) (26,16.58525599) (27,17.87364837) (28,17.1743113) (29,16.44062589) (30,16.07727472) (31,16.73314776) (32,18.79160489) (33,19.95891232) (34,18.07359767) (35,18.33409867) (36,17.68507042) (37,20.11109491) (38,18.25062226) (39,17.35011908) (40,18.3853611) (41,16.21627221) (42,17.8561945) (43,16.96257975) (44,17.71240609) (45,18.69398713) (46,16.60517565) (47,21.04931214) (48,20.43849543) (49,22.47462539) (50,18.46488472) (51,19.18485356) (52,21.34616905) (53,17.74280941) (54,19.85160266) (55,17.87933196) (56,17.67446815) (57,17.72410079) (58,15.48265042) (59,20.65613066) (60,22.13090232) (61,18.59352972) (62,22.81074009) (63,18.547482) (64,19.62291155) (65,22.98818331) (66,19.90146944) (67,20.05791639) (68,21.29805194) (69,21.16544559) (70,20.22457878) (71,23.60359895) (72,21.8813739) (73,23.47157971) (74,19.11512495) (75,20.07890132) (76,20.61678125) (77,20.74902565) (78,21.35512308) (79,22.98084298) (80,23.41986711) (81,22.05883007) (82,23.6224736) (83,22.41296439) (84,21.37809002) (85,22.10074991) (86,23.47331199) (87,24.39138999) (88,22.14136159) (89,23.1808487) (90,22.81106635) (91,23.35891367) (92,21.96949227) (93,18.78388389) (94,21.95333503) (95,23.33350207) (96,24.71716887) (97,21.29123648) (98,24.22398894) (99,24.57368232) (100,21.52236218) (101,23.26761754) (102,24.64532118) (103,22.74849036) (104,24.26677382) (105,23.55973685) (106,22.34270896)
    };
    \addlegendentry{{TR-TS}}

\end{axis}
\end{tikzpicture}
\caption{Analysis of linear vs. radix tree RZTs and the use of the timestamp technique to reduce wasteful lookups. The horizontal axis lists individual $19 \times 19$ problems as in Table \ref{tab:exp_data_19}. The vertical axis shows the traversing costs in log scale (base 2) of lookups.}
\label{fig:traverse_cost}
\end{figure}
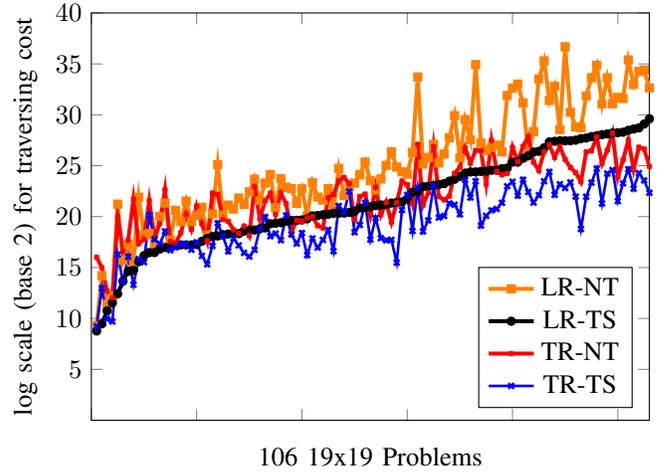

To illustrate the benefit of the radix tree data structure used to store and lookup table entries, we also implemented a linear lookup zone pattern table for comparison.
Both linear and radix tree RZTs, denoted by LR and TR respectively, were then used to solve L\&D problems in this experiment.
We measure the overhead of the RZT by the \emph{traversing cost}, which is the sum of the total number of table node visits and linked list access counts at the leaf nodes of the table.
Fig. \ref{fig:traverse_cost} shows the traversing costs on the 106 $19 \times 19$ problems in log scale (base 2) for the four settings LR-NT, LR-TS, TR-NT, TR-TS, where TS stands for using the timestamp technique to reduce wasteful tree lookups (subsection \ref{subsec:traverse_cost}), and NT otherwise. 
The traversing cost is measured for all 106 $19 \times 19$ L\&D problems, sorted by ascending order according to cost for LR-TS.
The results show that LR-TS outperforms others (even for the radix tree RZTs) for those with lower costs, about 20 problems in the leftmost part of Fig. \ref{fig:traverse_cost}, since its overhead is cheaper than maintaining a radix tree.
However, as the problems become more difficult and the number of lookups increases, the benefit of the tree structure begins to show.
In summary, the methods with the radix tree, namely TR-NT and TR-TS, performs better in large size problems. 
In addition, the methods with the timestamp mechanism (TR-TS and LR-TS) generally performs better than those (TR-NT and LR-NT) without it. 
Namely, the traversing cost of TR-TS is 9.31 times lower than TR-NT, or a reduction of about 89.26\% of the cost, i.e., nearly reduce the overhead to 10\% of the orginal cost.

One detail worth noting for the issue of many-to-many matching relation is that we return the first match during lookup.
However, in the context of RZS, matching smaller zone patterns leads to more move pruning for a more efficient search.
We tried finding all matching patterns and picking the entry with the smallest zone pattern and found that only 2.47\% of the cases resulted in a smaller zone pattern. 
The simulation count did not benefit significantly from the smaller pattern, while lookup cost increased by 20\%.
This leads us to conclude that returning the first hit during RZT lookup is the better choice.

\subsection{Empirical results for lookup time complexity}
\label{subsec:case_study}

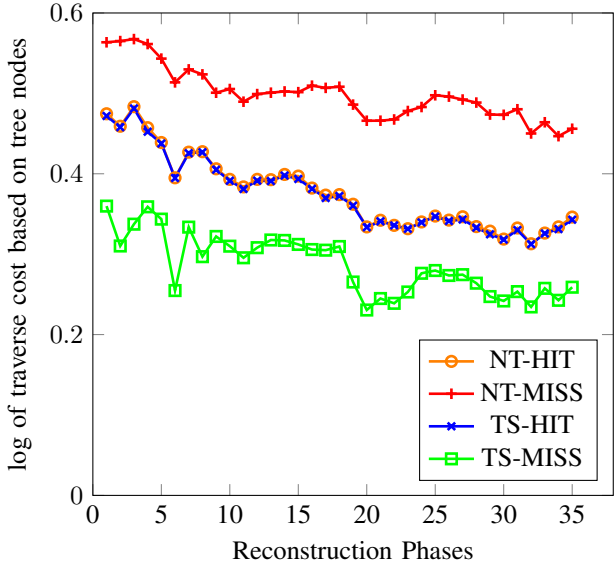
\begin{figure}[t]
\centering
\begin{tikzpicture}
\begin{axis}[
    height=8cm,
    width=8.5cm,
    xlabel={Reconstruction Phases},
    ylabel={log of traverse cost based on tree nodes},
    xmin=0, xmax=38,
    ymin=0, ymax=0.6,
    xtick={0,5,10,15,20,25,30,35},
    xticklabels={0,5,10,15,20,25,30,35},
    ytick={0, 0.2, 0.4, 0.6},
    legend pos=south east,
]

\addplot[
    line width=1pt,
    color=orange,
    mark=o,
    mark size=2pt,
    error bars/.cd,
    y dir = both,
    y explicit
    ]
    coordinates {
    (1,0.474643027) (2,0.4591335546) (3,0.4834817525) (4,0.4573563692) (5,0.4388788524) (6,0.3950321948) (7,0.4268779919) (8,0.4274829832) (9,0.4062103422) (10,0.3930660639) (11,0.383678855) (12,0.3930694756) (13,0.3926605919) (14,0.3991849866) (15,0.3971139558) (16,0.3822314817) (17,0.3734597865) (18,0.3741001283) (19,0.3621170236) (20,0.3339200715) (21,0.3421843687) (22,0.3358553473) (23,0.3317001642) (24,0.3400938425) (25,0.3474168522) (26,0.3423926063) (27,0.3465444729) (28,0.334310181) (29,0.328689446) (30,0.318781418) (31,0.3325822546) (32,0.3129771854) (33,0.3262716484) (34,0.3339722173) (35,0.3461431776)
    };
    \legend{NT-HIT}
    
\addplot[
    line width=1pt,
    color=red,
    mark=+,
    mark size=2pt,
    error bars/.cd,
    y dir = both,
    y explicit
    ]
    coordinates {
    (1,0.5635400321) (2,0.5650447633) (3,0.5675292153) (4,0.5613549432) (5,0.543245293) (6,0.5137010183) (7,0.5298303913) (8,0.5236424738) (9,0.5008268481) (10,0.5054413838) (11,0.4896757182) (12,0.4991347431) (13,0.5008735219) (14,0.5024042463) (15,0.5015222716) (16,0.5095797019) (17,0.5068452518) (18,0.5082685856) (19,0.4860367884) (20,0.4659953887) (21,0.4660285393) (22,0.4675561886) (23,0.478084473) (24,0.4830370919) (25,0.4976320577) (26,0.4959806943) (27,0.492350435) (28,0.4885646648) (29,0.4736683531) (30,0.4733457479) (31,0.4802591544) (32,0.4501475922) (33,0.4640804209) (34,0.4467916142) (35,0.456018492)
    };
    \addlegendentry{NT-MISS}
    
\addplot[
    line width=1pt,
    color=blue,
    mark=x,
    mark size=2pt,
    error bars/.cd,
    y dir = both,
    y explicit
    ]
    coordinates {
    (1,0.4719128345) (2,0.4579153354) (3,0.4812285603) (4,0.4523183871) (5,0.4376546867) (6,0.3947125529) (7,0.4255655715) (8,0.4268331643) (9,0.405028925) (10,0.3912827786) (11,0.3811203088) (12,0.3914264123) (13,0.3909395488) (14,0.3977945512) (15,0.3935733714) (16,0.3812405738) (17,0.369870976) (18,0.3723077394) (19,0.3603198093) (20,0.3334987335) (21,0.3409060171) (22,0.3358030829) (23,0.3314759696) (24,0.3395054522) (25,0.3468649369) (26,0.3413113109) (27,0.3430988029) (28,0.3330152347) (29,0.3247603787) (30,0.3181312988) (31,0.3300094167) (32,0.3124495762) (33,0.3257851495) (34,0.3312205022) (35,0.3428896731)
    };
    \addlegendentry{TS-HIT}

\addplot[
    line width=1pt,
    color=green,
    mark=square,
    mark size=2pt,
    error bars/.cd,
    y dir = both,
    y explicit
    ]
    coordinates {
    (1,0.359796039) (2,0.3102012002) (3,0.3373250067) (4,0.3589261959) (5,0.3435962292) (6,0.2547728879) (7,0.3336402525) (8,0.2969166351) (9,0.3220490537) (10,0.3099684534) (11,0.2955327016) (12,0.3080872617) (13,0.3176557468) (14,0.3171095217) (15,0.3121434485) (16,0.3057802368) (17,0.3049860695) (18,0.309542862) (19,0.2653751541) (20,0.2305638996) (21,0.2448754291) (22,0.2388745177) (23,0.2529838171) (24,0.2762446314) (25,0.2796894489) (26,0.2735175514) (27,0.2745370283) (28,0.2639388352) (29,0.2473125885) (30,0.2418406782) (31,0.2534494598) (32,0.2344102977) (33,0.2575387075) (34,0.2426501899) (35,0.2589237372)
    };
    \addlegendentry{TS-MISS}
\end{axis}
\end{tikzpicture}
\caption{The statistics of the zone tables on the $7 \times 7$ problem of ID 3.}
\label{fig:tt_statistic_3}
\end{figure}

To complement the theoretical discussion (subsection \ref{subsec:complexity}) of using a radix tree for the RZT with empirical results, we collect statistics for the three $7 \times 7$ problems that stored over 10,000 entries.
Let $n$ be the number of nodes in radix trees and $c$ be the traversing cost without counting the traversal in linked lists, in the radix tree of RZT. 
For the three  $7 \times 7$ problems, we count their averaged number of nodes $n_{av}$ and averaged traversing costs $c_{av}$ during each reconstruction phase, respectively shown in Figs. \ref{fig:tt_statistic_3}, \ref{fig:tt_statistic_10}, and \ref{fig:tt_statistic_12}.
The horizontal axis is the number of reconstructions to the radix tree, and the vertical axis is 
$(\log c_{av}) / (\log n_{av})$. 
Since the traversing cost $c$ must be smaller than $n$, the vertical axis must be smaller than one and is 0.5 if $c=\sqrt{n}$. 
The orange and red lines, respectively denoted by NT-HIT and NT-MISS, show costs for the cases where a hit or a miss occurred without using timestamps (NT), respectively.
The blue and green lines, respectively denoted by TS-HIT and TS-MISS, show costs for hits and misses with timestamp (TS), respectively.

With an observed hit rate of about 2\%\footnote{The hit rate, about 2\%, is derived based on the average ratio of hit counts against lookup counts, respectively shown in Tables \ref{tab:exp_data_7} and \ref{tab:exp_data_19}}, miss lookups dominate the performance. 
With the help of timestamp, the vast majority of lookups miss early in the top layers of the radix tree in most cases, and therefore TS-MISS has the greatest performance at about 0.2 from these figures. 
Both TS-HIT and NT-HIT are almost the same, since during hits, the lookup traverses to a leaf node then finds a matching pattern in the linked list whether timestamps are used or not.
Without timestamp, NT-MISS performs much worse since it always traverses through all possibilities, resulting in a larger cost.
We also observe that when the number of entries grows, the values in the vertical axis do not grow significantly, and get smaller even in some cases.

\begin{figure}[t]
\centering
\begin{tikzpicture}
\begin{axis}[
    height=8.2cm,
    width=8.5cm,
    xlabel={Reconstruction Phases},
    ylabel={log of traverse cost based on tree nodes},
    xmin=0, xmax=38,
    ymin=0, ymax=0.6,
    xtick={0,5,10,15,20,25,30,35},
    xticklabels={0,5,10,15,20,25,30,35},
    ytick={0, 0.2, 0.4, 0.6},
    legend pos=south east,
]

\addplot[
    line width=1pt,
    color=orange,
    mark=o,
    mark size=2pt,
    error bars/.cd,
    y dir = both,
    y explicit
    ]
    coordinates {
    (1,0.4780261319) (2,0.4428299081) (3,0.421773122) (4,0.4230814952) (5,0.4198276979) (6,0.3858746293) (7,0.3779467533) (8,0.3831816073) (9,0.3706485657) (10,0.3747107198) (11,0.3718190728) (12,0.3714676131) (13,0.3760020165) (14,0.3918756732) (15,0.3743819998) (16,0.3441394354) (17,0.3353879858) (18,0.3607858238) (19,0.3384308184) (20,0.3377848855) (21,0.3335565637) (22,0.3230344114) (23,0.3312836713) (24,0.3177966162) (25,0.3302931781) (26,0.3259031883) (27,0.3303695319) (28,0.3212783871) (29,0.3260870919) (30,0.3071908313) (31,0.3256587891) (32,0.3164362083) (33,0.3232514656) (34,0.3318663426) (35,0.3
   331444828)
    };
    \legend{NT-HIT}
    
\addplot[
    line width=1pt,
    color=red,
    mark=+,
    mark size=2pt,
    error bars/.cd,
    y dir = both,
    y explicit
    ]
    coordinates {
    (1,0.5146771065) (2,0.5067369046) (3,0.4914740031) (4,0.5116233905) (5,0.492873164) (6,0.4623385615) (7,0.4428914064) (8,0.4690492881) (9,0.439220075) (10,0.4645099906) (11,0.4285428376) (12,0.4178535371) (13,0.3895137873) (14,0.4712901946) (15,0.4753926644) (16,0.472131367) (17,0.4565226607) (18,0.4540466091) (19,0.4613633738) (20,0.4630674177) (21,0.4512213385) (22,0.4443100249) (23,0.4160974661) (24,0.4184952252) (25,0.4212620524) (26,0.4187038428) (27,0.4159409963) (28,0.4286763345) (29,0.451195373) (30,0.4305046263) (31,0.4293706681) (32,0.434690359) (33,0.4167966067) (34,0.4237842751) (35,0.4292502286)
    };
    \addlegendentry{NT-MISS}
    
\addplot[
    line width=1pt,
    color=blue,
    mark=x,
    mark size=2pt,
    error bars/.cd,
    y dir = both,
    y explicit
    ]
    coordinates {
    (1,0.4776432908) (2,0.4428299081) (3,0.4214970311) (4,0.4227176364) (5,0.4194716828) (6,0.3858746293) (7,0.3779467533) (8,0.3824728867) (9,0.3706485657) (10,0.3725211229) (11,0.3710926434) (12,0.3714676131) (13,0.3748195704) (14,0.3898947735) (15,0.3725490756) (16,0.3428200564) (17,0.3349123768) (18,0.3598388959) (19,0.3373828751) (20,0.3366662679) (21,0.3321717127) (22,0.3230234975) (23,0.3306308674) (24,0.3175855557) (25,0.3290468168) (26,0.3254253965) (27,0.3294181542) (28,0.3206086822) (29,0.3251964339) (30,0.3070325383) (31,0.3253217432) (32,0.3162177924) (33,0.3218824311) (34,0.3365644533) (35,0.3339597733)
    };
    \addlegendentry{TS-HIT}

\addplot[
    line width=1pt,
    color=green,
    mark=square,
    mark size=2pt,
    error bars/.cd,
    y dir = both,
    y explicit
    ]
    coordinates {
    (1,0.2413104957) (2,0.2464256629) (3,0.2422071294) (4,0.2448766388) (5,0.2516677391) (6,0.2427906676) (7,0.2220315876) (8,0.2112472972) (9,0.1904846309) (10,0.2081020396) (11,0.2242338888) (12,0.1810152814) (13,0.1980774274) (14,0.2345136199) (15,0.2216611695) (16,0.2075776922) (17,0.2099665486) (18,0.2225512373) (19,0.2334679277) (20,0.2453219611) (21,0.2294586813) (22,0.2223992155) (23,0.2182222989) (24,0.2306434979) (25,0.2375832062) (26,0.2279352113) (27,0.2140091047) (28,0.2191391911) (29,0.209615829) (30,0.200396319) (31,0.2050853062) (32,0.2184737989) (33,0.225325593) (34,0.2388750922) (35,0.2305387315)
    };
    \addlegendentry{TS-MISS}

\end{axis}
\end{tikzpicture}
\caption{The statistics of the zone tables on the $7 \times 7$ problem of ID 10.}
\label{fig:tt_statistic_10}
\end{figure}

\begin{figure}[t]
\centering
\begin{tikzpicture}
\begin{axis}[
    height=8cm,
    width=8.5cm,
    xlabel={Reconstruction Phases},
    ylabel={log of traverse cost based on tree nodes},
    xmin=0, xmax=39,
    ymin=0, ymax=0.6,
    xtick={0,5,10,15,20,25,30,35},
    xticklabels={0,5,10,15,20,25,30,35},
    ytick={0, 0.2, 0.4, 0.6},
    legend pos=south east,
]

\addplot[
    line width=1pt,
    color=orange,
    mark=o,
    mark size=2pt,
    error bars/.cd,
    y dir = both,
    y explicit
    ]
    coordinates {
    (1,0.4735814642) (2,0.4457897746) (3,0.4451613304) (4,0.4306794516) (5,0.4509509916) (6,0.4170749316) (7,0.4114678227) (8,0.4079530059) (9,0.3870162704) (10,0.3804731997) (11,0.3965050946) (12,0.4008031582) (13,0.3973032695) (14,0.3761621662) (15,0.3567399758) (16,0.3849320311) (17,0.3536776092) (18,0.3629172458) (19,0.3578408489) (20,0.3622118274) (21,0.358999957) (22,0.3593516987) (23,0.3451726261) (24,0.3581576166) (25,0.3548763387) (26,0.3601904665) (27,0.3425508824) (28,0.346539256) (29,0.3290072993) (30,0.32996055) (31,0.3303092151) (32,0.3200768641) (33,0.3216335992) (34,0.3145486225) (35,0.3314637139) (36,0.3201251596) (37,0.3275399938) (38,0.3361365974)
    };
    \legend{NT-HIT}
    
\addplot[
    line width=1pt,
    color=red,
    mark=+,
    mark size=2pt,
    error bars/.cd,
    y dir = both,
    y explicit
    ]
    coordinates {
    (1,0.5832320913) (2,0.4835585763) (3,0.4753666463) (4,0.5060054779) (5,0.5439585316) (6,0.5035631751) (7,0.5054864833) (8,0.4868495264) (9,0.4885145001) (10,0.4890665531) (11,0.4910122912) (12,0.4688381609) (13,0.4393718265) (14,0.45034445) (15,0.4360977046) (16,0.4492092679) (17,0.4245581372) (18,0.4519958352) (19,0.4524058494) (20,0.4359835328) (21,0.4354281558) (22,0.4541243025) (23,0.4447349676) (24,0.4410781198) (25,0.4462174855) (26,0.4495358382) (27,0.4693381819) (28,0.4757237126) (29,0.4753703018) (30,0.4831178845) (31,0.4910975689) (32,0.4724565197) (33,0.4767470441) (34,0.4705952595) (35,0.4735868006) (36,0.4687824442) (37,0.4380905201) (38,0.4392784028)
    };
    \addlegendentry{NT-MISS}
    
\addplot[
    line width=1pt,
    color=blue,
    mark=x,
    mark size=2pt,
    error bars/.cd,
    y dir = both,
    y explicit
    ]
    coordinates {
    (1,0.4733704127) (2,0.4457897746) (3,0.4433946213) (4,0.4303047296) (5,0.4479114674) (6,0.4157099201) (7,0.4091857529) (8,0.406989023) (9,0.3862459029) (10,0.3804731997) (11,0.3960266447) (12,0.396031861) (13,0.3971749101) (14,0.3758059339) (15,0.35637694) (16,0.3843648399) (17,0.3536776092) (18,0.3622800706) (19,0.3575401363) (20,0.3600303376) (21,0.3586310426) (22,0.3585162637) (23,0.3439823644) (24,0.3570634659) (25,0.3542615309) (26,0.3591868361) (27,0.3412267621) (28,0.3456346757) (29,0.3285169765) (30,0.3298210054) (31,0.3300726479) (32,0.3198180643) (33,0.3212617735) (34,0.3143673798) (35,0.3311924109) (36,0.3195930198) (37,0.3257607975) (38,0.3339159349)
    };
    \addlegendentry{TS-HIT}

\addplot[
    line width=1pt,
    color=green,
    mark=square,
    mark size=2pt,
    error bars/.cd,
    y dir = both,
    y explicit
    ]
    coordinates {
    (1,0.3189735678) (2,0.2307677015) (3,0.2341156078) (4,0.2716713069) (5,0.3172371793) (6,0.2982274243) (7,0.271789451) (8,0.2801010711) (9,0.239524231) (10,0.2464755555) (11,0.2681336237) (12,0.2720216649) (13,0.2389469341) (14,0.2393550907) (15,0.2297924656) (16,0.2360210217) (17,0.240361722) (18,0.2437438845) (19,0.2401947189) (20,0.2397293971) (21,0.2230593945) (22,0.236100012) (23,0.238880182) (24,0.2431854837) (25,0.2553758687) (26,0.2565449105) (27,0.2385019262) (28,0.24017965) (29,0.2342553962) (30,0.2517471062) (31,0.2429975522) (32,0.2491326794) (33,0.2512334799) (34,0.2442444554) (35,0.23871783) (36,0.2466660385) (37,0.2393876383) (38,0.2420409474)
    };
    \addlegendentry{TS-MISS}

\end{axis}
\end{tikzpicture}
\caption{The statistics of the zone tables on the $7 \times 7$ problem of ID 12.}
\label{fig:tt_statistic_12}
\end{figure}

\begin{table*}[b]
\centering
\caption{Data of $7 \times 7$ Problems} \label{tab:exp_data_7}
\begin{tabular}
{|c|c|c|c|c|c|c|c|c|c|}
\hline
 & \multicolumn{2}{c|}{Sim.} & \multicolumn{7}{c|}{RZT info.}\\
\hline
ID & w/o RZT & w/ RZT & Time & \#Entry & \#Lookup & LookupTime & \#Hit & \#Compare & \#Cost\\
\hline
1	&	325	&	277	&	0.3	&	60	&	1,489	&	0.002	&	44	&	8,737	&	10,226	\\	\hline
2	&	4,958	&	2,179	&	2.399	&	391	&	16,154	&	0.025	&	498	&	57,114	&	126,063	\\	\hline
3	&	217,461	&	79,448	&	119.783	&	10,088	&	1,072,269	&	3.381	&	16,816	&	3,198,853	&	16,652,668	\\	\hline
4	&	106,430	&	15,662	&	17.681	&	1,198	&	158,537	&	0.468	&	1,552	&	971,294	&	2,572,835	\\	\hline
5	&	204,691	&	19,331	&	26.906	&	3,740	&	237,909	&	0.429	&	5,139	&	113,680	&	1,778,601	\\	\hline
6	&	177,231	&	19,664	&	24.329	&	2,508	&	234,354	&	0.570	&	6,690	&	360,977	&	2,543,854	\\	\hline
7	&	173,056	&	24,445	&	37.858	&	4,647	&	330,709	&	0.960	&	6,814	&	302,711	&	4,427,764	\\	\hline
8	&	164,633	&	59,701	&	83.302	&	8,529	&	764,143	&	2.793	&	14,854	&	1,546,601	&	13,319,697	\\	\hline
9	&	180,901	&	29,342	&	43.275	&	4,267	&	359,269	&	1.834	&	8,323	&	694,944	&	8,683,018	\\	\hline
10	&	335,878	&	65,559	&	111.036	&	11,764	&	1,053,897	&	2.426	&	16,704	&	871,399	&	8,954,379	\\	\hline
11	&	449,791	&	93,456	&	142.028	&	8,409	&	1,418,016	&	3.829	&	9,661	&	2,471,584	&	16,881,650	\\	\hline
12	&	365,809	&	83,663	&	114.819	&	14,821	&	1,119,468	&	2.772	&	20,047	&	608,849	&	12,042,774	\\	\hline
13	&	1,129	&	521	&	0.431	&	94	&	3,562	&	0.003	&	63	&	19,577	&	23,139	\\	\hline
14	&	915	&	645	&	0.975	&	201	&	5,853	&	0.007	&	128	&	22,019	&	38,133	\\	\hline
15	&	2,402	&	972	&	0.894	&	325	&	6,969	&	0.010	&	216	&	26,886	&	52,363	\\	\hline
16	&	18,279	&	13,419	&	17.615	&	2,061	&	149,228	&	0.286	&	3,025	&	125,024	&	1,115,381	\\	\hline
17	&	319,276	&	52,828	&	81.91	&	6,247	&	720,778	&	2.253	&	13,756	&	1,827,587	&	10,523,026	\\	\hline
18	&	423,564	&	122,173	&	176.827	&	9,293	&	1,782,851	&	3.907	&	10,962	&	818,042	&	15,361,973	\\	\hline
19	&	7,067	&	4,027	&	4.361	&	791	&	33,167	&	0.053	&	666	&	39,989	&	200,402	\\	\hline
20	&	13,137	&	5,026	&	6.863	&	1,180	&	53,250	&	0.099	&	1,071	&	45,609	&	417,009	\\	\hline
\end{tabular}
\end{table*}

\begin{table*}[b]
\centering
\caption{Data of $19 \times 19$ Problems, the first column indicates ID, volume, and pages of the problem in the book \cite{cho1987life}. } \label{tab:exp_data_19}
\begin{tabular}
{|c|c|c|c|c|c|c|c|c|c|}
\hline
 & \multicolumn{2}{c|}{Sim.} & \multicolumn{7}{c|}{RZT info.}\\
\hline
ID-Vol.-P. & w/o RZT & w/ RZT & Time & \#Entry & \#Lookup & LookupTime & \#Hit & \#Compare & \#Cost\\
\hline
1-1-31	&	2,596	&	495	&	0.541	&	79	&	3,312	&	0.0028	&	30	&	12,273	&	3,312	\\	\hline
2-1-37	&	38,275	&	20,594	&	19.497	&	180	&	165,565	&	0.2118	&	516	&	1,412,281	&	174,304	\\	\hline
3-1-40	&	9,375	&	48,828	&	50.001	&	452	&	435,942	&	0.3281	&	518	&	1,047,945	&	520,509	\\	\hline
4-1-42	&	416	&	295	&	0.193	&	24	&	1,127	&	0.0005	&	7	&	977	&	1,127	\\	\hline
5-1-46	&	65,687	&	98,257	&	86.649	&	279	&	760,876	&	0.8804	&	3,060	&	1,186,585	&	2,256,747	\\	\hline
6-1-57	&	3,878	&	2,139	&	1.343	&	70	&	10,899	&	0.0101	&	154	&	68,384	&	10,899	\\	\hline
7-1-60	&	27,974	&	19,715	&	47.489	&	841	&	408,122	&	0.4996	&	1,180	&	1,157,247	&	1,420,948	\\	\hline
8-1-68	&	1,960	&	1,802	&	1.186	&	85	&	9,803	&	0.0066	&	152	&	40,761	&	9,803	\\	\hline
9-1-74	&	1,547	&	4,275	&	2.747	&	106	&	20,674	&	0.0251	&	536	&	221,875	&	20,942	\\	\hline
10-1-80	&	39,758	&	5,639	&	5.362	&	533	&	40,552	&	0.0520	&	703	&	137,104	&	140,722	\\	\hline
11-1-84	&	11,422	&	24,619	&	21.233	&	355	&	187,350	&	0.1088	&	700	&	234,010	&	265,218	\\	\hline
12-1-88	&	184,990	&	90,649	&	180.054	&	3,926	&	1,516,479	&	1.5167	&	7,327	&	308,623	&	3,933,357	\\	\hline
13-1-90	&	-	&	-	&	-	&	-	&	-	&	-	&	-	&	-	&	-	\\	\hline
14-1-98	&	-	&	-	&	-	&	-	&	-	&	-	&	-	&	-	&	-	\\	\hline
15-1-101	&	27,103	&	8,347	&	8.306	&	312	&	68,105	&	0.0382	&	330	&	122,766	&	93,244	\\	\hline
16-1-104	&	6,071	&	2,370	&	2.04	&	231	&	13,345	&	0.0171	&	230	&	118,571	&	23,968	\\	\hline
17-1-132	&	65,713	&	11,243	&	14.449	&	1,144	&	106,048	&	0.1043	&	1,638	&	226,453	&	275,153	\\	\hline
18-1-135	&	212,209	&	27,710	&	25.048	&	1,283	&	195,224	&	0.2339	&	3,699	&	714,119	&	572,632	\\	\hline
19-1-139	&	15,964	&	9,425	&	9.78	&	843	&	76,187	&	0.0635	&	1,285	&	82,551	&	202,828	\\	\hline
20-1-152	&	533,819	&	132,417	&	198.929	&	13,518	&	1,573,329	&	2.2740	&	29,879	&	432,244	&	8,187,658	\\	\hline
21-1-156	&	2,031	&	3,102	&	2.399	&	119	&	19,201	&	0.0102	&	84	&	45,894	&	20,452	\\	\hline
22-1-174	&	30,897	&	3,428	&	2.194	&	96	&	19,102	&	0.0142	&	410	&	99,691	&	19,102	\\	\hline
23-1-176	&	26,562	&	3,019	&	2.801	&	355	&	19,497	&	0.0176	&	473	&	43,014	&	45,625	\\	\hline
24-1-186	&	260,858	&	45,317	&	101.156	&	4,199	&	821,062	&	0.8574	&	9,375	&	456,203	&	2,716,841	\\	\hline
25-1-189	&	25,147	&	24,077	&	18.647	&	246	&	166,713	&	0.1469	&	599	&	882,352	&	230,441	\\	\hline
26-1-195	&	7,055	&	27,876	&	28.242	&	468	&	243,502	&	0.3199	&	791	&	137,391	&	1,125,245	\\	\hline
27-1-196	&	23,738	&	21,024	&	20.644	&	199	&	184,955	&	0.1808	&	422	&	999,309	&	323,944	\\	\hline
28-1-197	&	44,385	&	6,580	&	5.399	&	148	&	45,343	&	0.0517	&	520	&	364,219	&	62,798	\\	\hline
29-1-198	&	11,921	&	5,352	&	3.75	&	136	&	29,766	&	0.0235	&	206	&	138,370	&	33,929	\\	\hline
30-1-222	&	450,401	&	50,212	&	73.604	&	5,394	&	575,036	&	0.8184	&	9,361	&	270,546	&	2,870,751	\\	\hline
31-1-243	&	2,108	&	284	&	0.155	&	23	&	881	&	0.0003	&	5	&	829	&	881	\\	\hline
32-1-244	&	48,851	&	10,452	&	8.848	&	216	&	76,453	&	0.0893	&	525	&	148,682	&	258,227	\\	\hline
33-1-252	&	5,171	&	76,411	&	63.63	&	315	&	576,495	&	0.6434	&	2,323	&	1,507,906	&	1,734,448	\\	\hline
34-1-258	&	434	&	355	&	0.226	&	18	&	1,244	&	0.0006	&	19	&	595	&	1,244	\\	\hline
35-1-262	&	739	&	687	&	0.467	&	53	&	3,080	&	0.0021	&	32	&	7,970	&	3,080	\\	\hline
36-1-264	&	6,062	&	1,687	&	1.397	&	120	&	9,622	&	0.0105	&	403	&	75,435	&	10,322	\\	\hline
37-1-268	&	16,438	&	2,694	&	1.785	&	91	&	14,981	&	0.0094	&	268	&	58,470	&	14,981	\\	\hline
38-1-269	&	5,625	&	1,787	&	1.512	&	124	&	10,293	&	0.0119	&	407	&	97,415	&	11,202	\\	\hline
39-1-270	&	22,998	&	4,481	&	3.204	&	143	&	26,234	&	0.0216	&	232	&	154,931	&	29,874	\\	\hline
40-1-272	&	22,058	&	460	&	0.252	&	46	&	1,729	&	0.0015	&	25	&	10,183	&	1,729	\\	\hline
41-1-279	&	13,240	&	11,953	&	8.219	&	155	&	70,255	&	0.0669	&	240	&	447,962	&	76,064	\\	\hline
42-1-288	&	35,664	&	16,650	&	14.235	&	358	&	124,427	&	0.0686	&	659	&	225,599	&	174,692	\\	\hline
43-1-298	&	10,804	&	7,907	&	6.586	&	469	&	50,490	&	0.0496	&	801	&	125,176	&	140,080	\\	\hline
44-1-299	&	32,555	&	8,155	&	7.079	&	652	&	52,737	&	0.0595	&	609	&	89,730	&	179,521	\\	\hline
45-1-300	&	46,271	&	20,852	&	20.494	&	561	&	177,286	&	0.1432	&	702	&	62,732	&	456,963	\\	\hline
46-1-301	&	18,811	&	4,871	&	4.167	&	305	&	29,630	&	0.0249	&	441	&	104,979	&	52,367	\\	\hline
47-1-302	&	500,940	&	88,811	&	116.615	&	9,076	&	941,854	&	1.1314	&	16,324	&	401,584	&	3,552,420	\\	\hline
48-1-318	&	190,704	&	49,843	&	43.206	&	1,382	&	376,783	&	0.3527	&	2,488	&	442,708	&	1,159,269	\\	\hline
49-1-323	&	12,544	&	4,551	&	3.466	&	257	&	24,691	&	0.0392	&	628	&	223,602	&	75,272	\\	\hline
50-1-324	&	26,933	&	9,266	&	9.365	&	285	&	80,483	&	0.0374	&	524	&	56,142	&	113,286	\\	\hline
51-2-120	&	28,647	&	8,192	&	6.708	&	176	&	57,707	&	0.0282	&	135	&	103,539	&	63,106	\\	\hline
52-2-124	&	-	&	-	&	-	&	-	&	-	&	-	&	-	&	-	&	-	\\	\hline
53-2-125	&	180,507	&	29,469	&	30.136	&	1,490	&	249,624	&	0.3331	&	3,957	&	253,306	&	1,103,838	\\	\hline
54-2-126	&	435,557	&	92,900	&	154.984	&	11,613	&	1,208,389	&	1.7323	&	24,358	&	531,591	&	5,995,744	\\	\hline
55-2-127	&	58,900	&	58,758	&	73.118	&	1,944	&	606,477	&	0.6091	&	4,106	&	1,679,213	&	1,505,677	\\	\hline
56-2-129	&	-	&	-	&	-	&	-	&	-	&	-	&	-	&	-	&	-	\\	\hline
57-2-132	&	-	&	-	&	-	&	-	&	-	&	-	&	-	&	-	&	-	\\	\hline
58-2-133	&	5,287	&	2,872	&	2.708	&	121	&	20,190	&	0.0172	&	551	&	106,199	&	23,938	\\	\hline
59-2-138	&	1,030,394	&	63,667	&	64.121	&	419	&	549,545	&	0.3670	&	3,966	&	1,332,431	&	825,053	\\	\hline
60-2-139	&	195,445	&	49,254	&	42.187	&	1,409	&	360,174	&	0.9230	&	3,263	&	7,671,460	&	1,008,894	\\	\hline
61-2-142	&	522,342	&	46,556	&	59.157	&	1,287	&	462,799	&	0.6374	&	3,973	&	499,669	&	1,725,427	\\	\hline
62-2-144	&	-	&	-	&	-	&	-	&	-	&	-	&	-	&	-	&	-	\\	\hline
63-2-146	&	1,393,546	&	196,592	&	220.352	&	5,932	&	1,852,020	&	3.0994	&	27,493	&	765,574	&	11,184,805	\\	\hline
64-2-147	&	251,136	&	22,992	&	24.75	&	793	&	163,618	&	0.2835	&	2,403	&	167,600	&	975,371	\\	\hline
65-2-152	&	-	&	-	&	-	&	-	&	-	&	-	&	-	&	-	&	-	\\	\hline
66-2-156	&	59,898	&	26,871	&	27.743	&	1,252	&	207,219	&	0.3769	&	2,200	&	516,486	&	1,343,132	\\	\hline
67-2-158	&	-	&	-	&	-	&	-	&	-	&	-	&	-	&	-	&	-	\\	\hline
68-2-162	&	373,086	&	12,566	&	20.625	&	783	&	174,757	&	0.1740	&	773	&	210,508	&	532,089	\\	\hline
\end{tabular}
\end{table*}

\begin{table*}[b]
\centering
\begin{tabular}
{|c|c|c|c|c|c|c|c|c|c|}
\hline
 & \multicolumn{2}{c|}{Sim.} & \multicolumn{7}{c|}{RZT info.}\\
\hline
ID-Vol.-P. & w/o RZT & w/ RZT & Time & \#Entry & \#Lookup & LookupTime & \#Hit & \#Compare & \#Cost\\
\hline
69-2-164	&	-	&	-	&	-	&	-	&	-	&	-	&	-	&	-	&	-	\\	\hline
70-2-166	&	-	&	-	&	-	&	-	&	-	&	-	&	-	&	-	&	-	\\	\hline
71-2-245	&	-	&	-	&	-	&	-	&	-	&	-	&	-	&	-	&	-	\\	\hline
72-2-246	&	5,448	&	7,931	&	6.916	&	192	&	58,463	&	0.0397	&	316	&	147,649	&	74,539	\\	\hline
73-2-248	&	-	&	-	&	-	&	-	&	-	&	-	&	-	&	-	&	-	\\	\hline
74-2-252	&	7,483	&	4,234	&	4.518	&	335	&	31,820	&	0.0402	&	545	&	86,718	&	128,032	\\	\hline
75-2-253	&	331,103	&	88,034	&	96.926	&	644	&	860,912	&	1.1017	&	2,248	&	4,383,139	&	2,062,148	\\	\hline
76-2-254	&	-	&	-	&	-	&	-	&	-	&	-	&	-	&	-	&	-	\\	\hline
77-2-257	&	-	&	-	&	-	&	-	&	-	&	-	&	-	&	-	&	-	\\	\hline
78-2-260	&	-	&	-	&	-	&	-	&	-	&	-	&	-	&	-	&	-	\\	\hline
79-2-262	&	-	&	-	&	-	&	-	&	-	&	-	&	-	&	-	&	-	\\	\hline
80-2-274	&	-	&	-	&	-	&	-	&	-	&	-	&	-	&	-	&	-	\\	\hline
81-2-326	&	-	&	-	&	-	&	-	&	-	&	-	&	-	&	-	&	-	\\	\hline
82-2-328	&	43,247	&	10,534	&	13.549	&	1,229	&	102,971	&	0.1021	&	1,341	&	193,969	&	292,134	\\	\hline
83-2-329	&	2,009	&	1,944	&	1.354	&	103	&	9,604	&	0.0121	&	316	&	106,651	&	9,604	\\	\hline
84-2-330	&	22,957	&	3,695	&	3.114	&	133	&	25,757	&	0.0126	&	77	&	43,528	&	28,016	\\	\hline
85-2-331	&	-	&	-	&	-	&	-	&	-	&	-	&	-	&	-	&	-	\\	\hline
86-2-332	&	596,695	&	93,224	&	201.951	&	7,211	&	1,516,231	&	4.9561	&	21,246	&	2,210,819	&	21,311,357	\\	\hline
87-2-333	&	9,701	&	5,053	&	3.667	&	285	&	26,173	&	0.0508	&	401	&	176,322	&	90,013	\\	\hline
88-2-334	&	905,238	&	19,649	&	20.489	&	345	&	142,788	&	0.2057	&	771	&	95,069	&	715,160	\\	\hline
89-2-336	&	15,996	&	25,255	&	22.066	&	599	&	191,914	&	0.1135	&	834	&	293,334	&	260,393	\\	\hline
90-2-337	&	10,544	&	2,956	&	2.063	&	113	&	15,443	&	0.0198	&	642	&	162,659	&	15,932	\\	\hline
91-2-338	&	33,088	&	107,099	&	93.381	&	606	&	827,943	&	0.8613	&	1,415	&	5,418,627	&	1,237,542	\\	\hline
92-2-340	&	-	&	-	&	-	&	-	&	-	&	-	&	-	&	-	&	-	\\	\hline
93-2-341	&	16,056	&	11,231	&	9.811	&	450	&	71,348	&	0.0793	&	935	&	70,144	&	242,314	\\	\hline
94-2-342	&	173,118	&	18,665	&	21.234	&	1,044	&	158,647	&	0.2048	&	2,255	&	350,823	&	615,217	\\	\hline
95-2-343	&	12,210	&	2,115	&	2.241	&	207	&	14,786	&	0.0169	&	702	&	38,669	&	48,546	\\	\hline
96-2-344	&	37,943	&	21,120	&	18.616	&	544	&	159,346	&	0.0965	&	1,076	&	122,462	&	256,224	\\	\hline
97-2-345	&	27,276	&	7,645	&	7.131	&	65	&	66,260	&	0.0268	&	81	&	78,750	&	66,260	\\	\hline
98-2-351	&	219,067	&	94,313	&	140.699	&	3,038	&	1,195,259	&	1.4507	&	5,921	&	338,496	&	4,963,304	\\	\hline
99-2-352	&	23,855	&	6,420	&	5.798	&	361	&	44,550	&	0.0376	&	493	&	116,732	&	94,891	\\	\hline
100-2-353	&	303,910	&	54,330	&	122.933	&	4,220	&	940,197	&	1.3840	&	10,001	&	884,230	&	4,116,774	\\	\hline
101-2-355	&	-	&	-	&	-	&	-	&	-	&	-	&	-	&	-	&	-	\\	\hline
102-2-359	&	-	&	-	&	-	&	-	&	-	&	-	&	-	&	-	&	-	\\	\hline
103-2-363	&	-	&	-	&	-	&	-	&	-	&	-	&	-	&	-	&	-	\\	\hline
104-2-364	&	17,458	&	6,219	&	10.243	&	661	&	74,554	&	0.1291	&	1,072	&	180,422	&	490,091	\\	\hline
105-2-366	&	59,295	&	21,795	&	20.521	&	755	&	177,572	&	0.1642	&	846	&	73,708	&	527,947	\\	\hline
106-2-372	&	165,019	&	19,293	&	20.817	&	730	&	160,931	&	0.1757	&	823	&	117,154	&	495,126	\\	\hline
\end{tabular}
\end{table*}

\section{Conclusion}
\label{sec:conclusion}
In this paper, we propose a zone pattern table that is stored as a radix tree. This table collects zone patterns that are automatically generated during search. 
When in use with RZS \cite{shih2022anovel}, matches in the zone pattern table can correctly be set to winning, making solving positions in Go and Kill-all Go more efficient.
Future directions include dealing with ko (or SSK) and GHI, shifting or rotating zone patterns while guaranteeing correctness, and investigating whether zone patterns can be used for interpreting neural networks.

\bibliography{bibliography}

\vspace{-8 mm}

\begin{IEEEbiography}[{\includegraphics[width=0.9in,height=1.1in,clip,keepaspectratio]{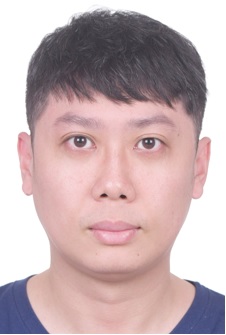}}]{Chung-Chin Shih}
is currently a Ph.D. candidate in the Department of Computer Science at National Yang Ming Chiao Tung University. His research interests include machine learning, reinforcement learning and computer games.
\end{IEEEbiography}

\vspace{-15 mm}

\begin{IEEEbiography}[{\includegraphics[width=1in,height=1.25in,clip,keepaspectratio]{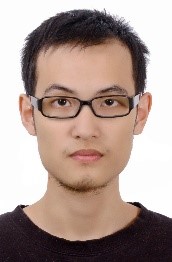}}]{Ting Han Wei}
is currently a postdoctoral fellow in the Department of Computing Science at University of Alberta. He received his Ph.D. in Computer Science from National Chiao Tung University in 2019. His research interests include artificial intelligence, computer games and grid computing. He also published several papers in top-tier conferences, such as AAAI, IJCAI and ICLR.
\end{IEEEbiography}

\vspace{-15 mm}

\begin{IEEEbiography}[{\includegraphics[width=1in,height=1.25in,clip,keepaspectratio]{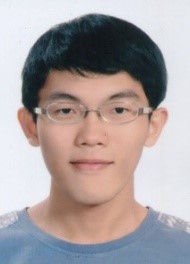}}]{Ti-Rong Wu}
is currently a assistant research fellow in the institute of Information Science at Academia Sinica. He received his Ph.D. in Computer Science from National Chiao Tung University in 2020. His research interests include machine learning, deep learning and computer games. He is the leader of the development team for the Go program, CGI, which won many competitions including the second place in the first World AI Open held in Ordos, China, 2017. He also published several papers in top-tier conferences, such as AAAI, ICLR, and NeurIPS. 
\end{IEEEbiography}

\vspace{-12 mm}

\begin{IEEEbiography}[{\includegraphics[width=1in,height=1.25in,clip,keepaspectratio]{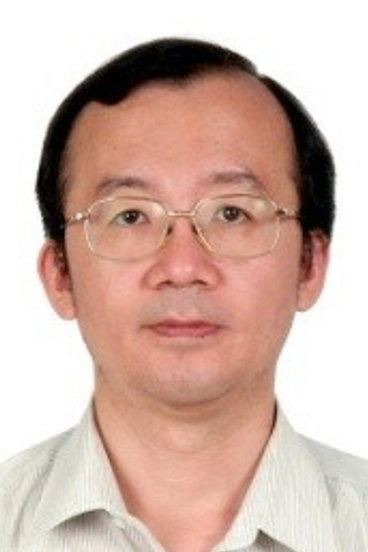}}]{I-Chen Wu}
(M’05-SM’15) is with the Department of Computer Science, at National Yang Ming Chiao Tung University. He received his B.S. in Electronic Engineering from National Taiwan University (NTU), M.S. in Computer Science from NTU, and Ph.D. in Computer Science from Carnegie-Mellon University, in 1982, 1984 and 1993, respectively. He serves in the editorial board of the IEEE Transactions on Games and ICGA Journal. His research interests include computer games and deep reinforcement learning. 
Dr. Wu introduced the new game, Connect6, a kind of six-in-a-row game. Since then, Connect6 has become a tournament item in Computer Olympiad. He led a team developing various game playing programs, winning over 60 gold medals in international tournaments, including Computer Olympiad. He wrote over 160 papers, and served as chairs and committee in over 50 academic conferences and organizations, including the conference chair of IEEE CIG conference 2015.
\end{IEEEbiography}

\end{document}